\documentclass[sigconf]{acmart}

\usepackage{multirow}
\usepackage{graphicx}
\usepackage{pifont}
\usepackage{balance}

\AtBeginDocument{%
  }

\copyrightyear{2025}
\acmYear{2025}
\setcopyright{cc}
\setcctype{by}
\acmConference[SVC '25]{Proceedings of the 1st International Workshop \& Challenge on Subtle Visual Computing}{October 27--28, 2025}{Dublin, Ireland}
\acmBooktitle{Proceedings of the 1st International Workshop \& Challenge on Subtle Visual Computing (SVC '25), October 27--28, 2025, Dublin, Ireland}\acmDOI{10.1145/3728425.3759909}
\acmISBN{979-8-4007-1837-3/2025/10}

\begin{document}

\title{MADPromptS: Unlocking Zero-Shot Morphing Attack Detection with Multiple Prompt Aggregation}


\author{Eduarda Caldeira}
\orcid{0009-0002-4891-0057}
\affiliation{%
  \institution{Fraunhofer IGD}
  \city{Darmstadt}
  \country{Germany}
}
\email{eduarda.caldeira@igd.fraunhofer.de}

\author{Fadi Boutros}
\orcid{0000-0003-4516-9128}
\affiliation{%
  \institution{Fraunhofer IGD}
  \city{Darmstadt}
  \country{Germany}
}
\email{fadi.boutros@igd.fraunhofer.de}

\author{Naser Damer}
\orcid{0000-0001-7910-7895}
\affiliation{%
  \institution{Fraunhofer IGD and Department of Computer Science, TU Darmstadt}
  \city{Darmstadt}
  \country{Germany}
}
\email{naser.damer@igd.fraunhofer.de}

\renewcommand{\shortauthors}{Eduarda Caldeira, Fadi Boutros and Naser Damer}

\begin{abstract}
  Face Morphing Attack Detection (MAD) is a critical challenge in face recognition security, where attackers can fool systems by interpolating the identity information of two or more individuals into a single face image, resulting in samples that can be verified as belonging to multiple identities by face recognition systems. While multimodal foundation models (FMs) like CLIP offer strong zero-shot capabilities by jointly modeling images and text, most prior works on FMs for biometric recognition have relied on fine-tuning for specific downstream tasks, neglecting their potential for direct, generalizable deployment. This work explores a pure zero-shot approach to MAD by leveraging CLIP without any additional training or fine-tuning, focusing instead on the design and aggregation of multiple textual prompts per class. By aggregating the embeddings of diverse prompts, we better align the model’s internal representations with the MAD task, capturing richer and more varied cues indicative of bona-fide or attack samples. Our results show that prompt aggregation substantially improves zero-shot detection performance, demonstrating the effectiveness of exploiting foundation models’ built-in multimodal knowledge through efficient prompt engineering. The code is publicly released: \url{https://github.com/EduardaCaldeira/MADPromptS}
\end{abstract}




\maketitle

\section{Introduction}

In recent years, the research community's strong focus on deep learning (DL) techniques enabled the high-paced development of high-performing systems in different domains, including face recognition (FR) \cite{boutros2022elasticface, deng2019arcface}. Despite the unquestionable benefits associated with this evolution, the same scientific advances capable of enhancing biometric systems can also be maliciously deployed to attack them \cite{damer2018morgan, DBLP:conf/icb/FerraraFM14}, raising concerns about their secure deployment. Some of these malicious samples are created by interpolating identity information of two or more individuals in a single face image, resulting in samples that can be verified as belonging to multiple identities by FR systems and by humans. These attacks are known as morphing attacks (MA) due to their inherent property of incorporating defining features from multiple identities, contributing to impaired FR functionality when left undetected \cite{damer2023mordiff, caldeira2023unveiling, DBLP:journals/tbbis/ZhangVRRDB21}. The inability to efficiently detect these samples is particularly problematic in high-security applications, as they potentiate crimes such as identity theft. To mitigate such risks, various morphing attack detection (MAD) systems have been proposed in the recent years \cite{huber2022syn, caldeira2023unveiling, fang2022unsupervised, damer2021pw, neto2022orthomad, ramachandra2019detecting, caldeira2025madation}. These algorithms aim at distinguishing MAs from authentic (bonafide) samples before they are fed to FR systems, removing malicious samples from the recognition framework at an early stage and preventing them from being considered for face verification.

Foundation models (FMs) are large-scale networks that can be trained with unlabeled data, following a self-supervised learning paradigm \cite{oquab2023dinov2, radford2021learning, kirillov2023segment}. This allows FMs to be trained in massive and diverse datasets, resulting in trained models that can efficiently generalize to a wide variety of tasks \cite{DBLP:journals/corr/abs-2108-07258}. Due to this property, FMs can be directly deployed for classification of samples belonging to categories that were not necessarily analyzed during their training stage (zero-shot learning), which makes them powerful tools in fields that address several tasks, such as natural language processing \cite{brown2020language} and computer vision \cite{kirillov2023segment, ravi2024sam, oquab2023dinov2, radford2021learning}. While FMs have shown significant zero-shot capacity across several downstream tasks, they achieve less optimal performance when applied to domain-specific settings \cite{ravi2024sam}, for which an adaption to the downstream task is usually performed \cite{chen2022adaptformer, chen2022vision, hu2021lora}. Despite allowing for a beneficial balance between pre-trained FMs in-built information and the acquisition of domain-specific knowledge \cite{caldeira2025madation, DBLP:conf/wacv/OzgurCCBRD25}, this adaption process requires model fine-tuning, resulting in decreased computational efficiency when compared with zero-shot learning, which does not require any MAD training data. Prompt engineering has recently gained attention as an effective strategy to boost the zero-shot performance of foundation models without any additional computational burden \cite{radford2021learning}. By feeding multimodal FMs with carefully designed textual prompts describing each class, it is possible to better align the input representations with the built-in knowledge of the FM and, consequently, take better advantage of the FMs zero-shot capacity. 

In this work, we explore the potential of prompt engineering for zero-shot learning MAD. In particular, we analyze how the utilization of multiple prompts to describe each possible output class impacts zero-shot performance, highlighting the benefits of a careful selection of the prompt sets used during inference. The obtained results show that while using a single prompt per class provides a general description of the desired label, using multiple prompts allows the incorporation of more specific characteristics, shifting the attention of the FM to a broader spectrum of details. While some sets of prompts do not contribute to increase zero-shot performance, disjoint sets that contribute positively to the FM's performance present complementary properties, with their joint utilization resulting in further MAD performance improvements. Hence, this study highlights the benefits that can arise from efficient prompt engineering strategies, providing important insights regarding the importance of correctly exploiting textual prompts to the user's advantage by leveraging FM’s zero-shot learning ability to a more complete extent.

\section{Related Work} \label{sec:sota}
This section presents an overview of recent works proposing MAD solutions, followed by a discussion of recent advances in foundation models and their applications within the biometric domain.
\subsection{Morphing Attack Detection}

Interpolating the identities of two or more face images in one image,  such that it can be verified as belonging to those identities \cite{caldeira2023unveiling} 
is a major risk to many processes involving automatic or manual identity verification \cite{ferrara2016effects, scherhag2017vulnerability}. Detecting such attacks became a major challenge given the realistic appearance and the ease of creation of such morphing samples, motivating the research community towards the development of more accurate and generalizable MAD solutions \cite{huber2022syn, caldeira2023unveiling, fang2022unsupervised, damer2021pw, neto2022orthomad, ramachandra2019detecting, zhang2024generalized}. 
From an operational point of view, MAD solutions can be categorized as single image MAD strategies \cite{huber2022syn, caldeira2023unveiling, fang2022unsupervised, damer2021pw, neto2022orthomad, ramachandra2019detecting, DBLP:conf/iwbf/IvanovskaS23, zhang2024generalized}, which base their decision solely on the inspected image, and differential MAD solutions \cite{damer2019detecting}, which consider a live captured image along with the inspected sample. The latter strategies generally show higher accuracy in detecting morphing attacks, since they have access to additional information to make a prediction \cite{damer2019detecting,DBLP:conf/wacv/RamachandraV025,DBLP:conf/wacv/RamachandraVDVG24}. However, their applicability is limited in several use cases, as it requires performing a live capture under operator supervision. Hence, several studies have developed single-image MAD systems \cite{huber2022syn, caldeira2023unveiling, fang2022unsupervised, damer2021pw, neto2022orthomad, ramachandra2019detecting, DBLP:conf/iwbf/IvanovskaS23, zhang2024generalized}, which can be applied without the need to perform a live capture, allowing MAD in a wider range of real-world applications (e.g. analyzing standalone documents).

Recent single-image MAD works have explored diverse methods to detect morphing attacks, ranging from handcrafted features \cite{ramachandra2019detecting} to advanced deep learning techniques \cite{zhang2024generalized, DBLP:conf/iwbf/IvanovskaS23, caldeira2025madation}. Ramachandra~\textit{et al.} \cite{ramachandra2019detecting} proposed to extract multi-scale textural descriptors and classify them using collaborative representation. Another work \cite{damer2021pw} suggested that each pixel (or block of pixels) should be individually classified as bona-fide or morphing attack, shifting away from the common global classification towards a more localized detection. Unsupervised \cite{fang2022unsupervised} and self-supervised \cite{DBLP:conf/iwbf/IvanovskaS23} approaches have also been proposed to tackle the MAD task. In \cite{fang2022unsupervised}, the authors trained a robust autoencoder for anomaly detection using an unsupervised self-paced learning approach. This approach identifies suspicious training samples and assigns them less importance during training, despite the datasets being polluted with morphed samples. \cite{DBLP:conf/iwbf/IvanovskaS23} trained a self-supervised diffusion model to reconstruct bona-fide samples. While authentic samples can be easily reconstructed by this model, its ability to reconstruct morphed samples is significantly lower which results in higher reconstruction errors, allowing for detecting these malicious samples. \cite{huber2022syn} promoted the SYN-MAD 2022 competition on MAD based on synthetic training data, presenting a comprehensive analysis of the results of seven submitted approaches. In \cite{neto2022orthomad, caldeira2023unveiling}, morphing attacks were detected through the identification of independent identity information in each analyzed sample. Neto~\textit{et al.} \cite{neto2022orthomad} used orthogonal vectors to identify the presence of more than one identity in the input samples. \cite{caldeira2023unveiling} used knowledge distillation to transfer information from an auto-encoder trained on bona-fide samples, following distinct distillation techniques for bona-fide images (single identity distillation) and morphing attacks (double identity distillation).
A vision transformer architecture for MAD was presented in \cite{zhang2024generalized}, showing promising results. Very recently, MADation \cite{caldeira2025madation} used LoRa layers to fine-tune a foundation model to the downstream MAD task, highlighting the potential of FM in domain-specific tasks such as MAD. The work proposed in this document also focuses on single-image MAD due to its wider utility in real-world scenarios where a live probe might not be available. 

\subsection{Foundation Models}

Foundation models are large-scale networks that can be trained with unlabeled data, following a self-supervised learning paradigm. This allows FMs to be trained on massive and diverse datasets, resulting in trained models that can efficiently generalize to a wide variety of tasks \cite{DBLP:journals/corr/abs-2108-07258}. The DINOv2 family of networks \cite{oquab2023dinov2} comprises self-supervised visual models capable of producing universal features that can be applied to both image-level and pixel-level tasks. The Segment Anything Model (SAM) \cite{kirillov2023segment} demonstrates strong generalization capabilities, enabling zero-shot image segmentation across a wide range of domains. Contrastive Language-Image Pretraining (CLIP) \cite{radford2021learning} is a multimodal FM constituted by a visual encoder and a text encoder. This allows CLIP to consider visual and textual information simultaneously during its training process and effectively learn the correlation between images and their textual description. 

Recent advances in FMs have led the scientific community to explore their applicability to a wide range of downstream tasks, including in biometrics \cite{DBLP:journals/corr/abs-2108-07258, shahreza2025foundation}. One of the first works that applied foundation models to the face recognition task \cite{chettaoui2024froundation} concluded that using LoRA layers \cite{hu2021lora} to fine-tune FMs to the FR downstream task consistently outperforms training those models from scratch in low data availability scenarios. MADation \cite{caldeira2025madation} highlighted the adaptability of CLIP \cite{radford2021learning} to the domain-specific MAD task by fine-tuning its image encoder with LoRA layers. Similarly, FoundPAD \cite{DBLP:conf/wacv/OzgurCCBRD25} leverages LoRA-adapted CLIP and a binary classifier for face presentation attack detection (PAD), achieving superior performance across cross-dataset evaluations compared to state-of-the-art (SOTA) PAD systems. These results highlight the high adaptability of FMs to unseen downstream tasks even in domain-specific scenarios such as MAD and PAD. Arc2Face \cite{papantoniou2024arc2face} adopts foundation models to extract identity embeddings from face images, using these as conditions for a diffusion model to generate identity-specific synthetic faces. CLIB-FIQA \cite{DBLP:conf/cvpr/OuL0K24} used CLIP to perform face image quality assessment by aligning the visual features of input face images with textual descriptions of image quality factors such as pose and expression.

Apart from visual encoders, recent developments in large language models (LLMs) and, in particular, multimodal versions of GPT-4, have opened new research directions in the biometrics field. Hassanpour et al. \cite{DBLP:conf/icip/HassanpourKOYM24} assessed GPT-4's performance in tasks such as FR, gender classification, and age estimation. DeAndres-Tame et al. \cite{DBLP:journals/access/DeAndresTameTVMFO24} conducted a thorough evaluation of GPT-4 for face verification and soft-biometric attribute estimation, providing a complementary explainability analysis of the model's decisions. Despite the absence of fine-tuning to biometric tasks, GPT-4 managed to achieve 94\% verification accuracy on the LFW dataset \cite{huang2008labeled} and 96.3\% gender classification accuracy on MAAD-Face \cite{DBLP:journals/tifs/TerhorstFKDKK21}. Furthermore, GPT-4's capacity to match human faces was proved to be on par with average human performance \cite{kramer2025face}. Farmanifard and Ross \cite{DBLP:conf/icb/FarmanifardR24} explored GPT-4’s capabilities in iris recognition under zero-shot settings.

Recent works have explored the use of FM for biometric tasks, including MAD \cite{caldeira2025madation}, primarily by fine-tuning these models on task-specific datasets. While fine-tuning FM can yield strong performance on detecting MAs \cite{caldeira2025madation}, this approach often overlooks one of the key strengths of FMs, which is their ability to perform zero-shot predictions on previously unseen scenarios. By relying heavily on fine-tuning \cite{caldeira2025madation} or training models from scratch \cite{zhang2024generalized, DBLP:conf/iwbf/IvanovskaS23,neto2022orthomad}, these approaches may limit the generalization capacity of the models, which is especially critical in MAD applications where new morphing techniques continually emerge and the ability to detect unseen attacks without extensive retraining or fine-tuning is essential. This paper unleashes the power of multimodal FMs under a zero-shot prediction scenario, namely CLIP, for MAD by leveraging multiple prompts, aiming to better capture diverse cues of bona-fide and attack samples, and thus, improve the MAD performance.

\section{Zero-shot MAD using CLIP and multiple textual prompts aggregation} \label{sec:methodology}
This work leverages the zero-shot learning capability of the multimodal (text–image) foundation model CLIP for MAD by employing multiple carefully designed textual prompts per class (attack or bona-fide). By averaging these prompts, we aim to better align CLIP’s aggregated text embeddings with the image embeddings and, thus, capture more diverse cues without any need for fine-tuning. In this section, we first present preliminaries on CLIP, followed by a detailed description of the single- and multiple-prompt strategies for MAD.

\subsection{Preliminary on CLIP} \label{sec:clip}

CLIP \cite{radford2021learning} is a multimodal FM constituted by a visual encoder and a text encoder. CLIP was trained using a massive dataset where each image is paired with a textual description that may accurately describe it (positive pair) or not (negative pair), utilizing a contrastive learning paradigm where the cosine similarity between the features extracted for image and text is maximized/minimized for positive/negative pairs. This allows CLIP to learn the relationship between visual and textual inputs and simultaneously interpret them during inference, resulting in a model generalizable across distinct tasks \cite{radford2021learning} with very competitive zero-shot learning results. 

\subsection{MADPrompts}
In this work, we explore the zero-shot learning capacity of multimodal FMs in the MAD task when single and multiple prompts are used to describe each possible classification label. Using CLIP to classify images that are not present during the CLIP training process might result in suboptimal performance, particularly in domain-specific settings such as MAD \cite{caldeira2025madation} or face recognition \cite{chettaoui2024froundation}. However, works that highlight this phenomenon \cite{chettaoui2024froundation,caldeira2025madation,DBLP:conf/wacv/OzgurCCBRD25} only considered a single yet simple text prompt and do not provide extensive zero-shot evaluation using multiple prompts. Furthermore, they sometimes failed to adhere to the input image specifications that are expected to work optimally for CLIP. Hence, we propose to perform such an evaluation to reach a more comprehensive understanding of FMs' zero-shot capability and raise awareness towards the importance of efficient prompt engineering.

\subsubsection{Single-Prompt Inference}

When a multimodal FM is used to infer the label of a sample $x_i$ in a zero-shot learning setting, two parallel processing steps are required, one for each encoder. The visual encoder, $E_v$, processes $x_i$, producing an image embedding $e_i = E_v \,(x_i)$. The text encoder, $E_t$, processes a list of text prompts, each describing one of the possible labels of $x_i$, $y_i$. For a binary task such as MAD, where $y_i\in\{0,1\}$, $E_t$ is fed two individual text prompts describing bona-fide ($p_{BF}$) and morphing attack ($p_{MA}$) samples. This results in two text embeddings, $e_{BF} = E_t \,(p_{BF})$ and $e_{MA} = E_t \,(p_{MA})$, that represent the two possible labels in a feature space with the same dimensionality as $e_i$. It is important to note that all the considered embeddings are normalized, thus laying on a unit hypersphere with the same dimensionality. As explained in Section \ref{sec:clip}, the contrastive learning paradigm of multimodal FMs like CLIP results in a strong correlation between the feature space of the visual and text embeddings. Hence, the zero-shot learning prediction for $x_i$, $\hat{y}_i$, can be determined by selecting the label whose embedding is more similar to $e_i$:

\begin{equation}
\hat{y}_i = 
\begin{cases} 
    0, & \phi(e_i, e_{BF}) > \phi(e_i, e_{MA}) \\
    1, & \text{otherwise}
\end{cases}, \label{eq:predict_label}
\end{equation}
where $\phi(.)$ is the cosine similarity function.

\subsubsection{Prompts Aggregation}
Although CLIP can achieve remarkable zero-shot evaluation results in several tasks using a single text prompt to define each class, this model has been shown to perform better when the information of several text prompts defining each possible class is combined into a single text embedding \cite{radford2021learning}. The proposed approach combines multiple context prompts on the text feature space by averaging their contributions before comparing the final feature vector with the visual embedding, increasing CLIP's zero-shot performance on e.g., ImageNet by 3.5 percentage points \cite{radford2021learning}. These results suggest that customizing the prompts used to perform zero-shot classification can largely contribute to its success.

Taking this into consideration, we further adapt CLIP's zero-shot evaluation to include multiple text prompts per class. In this scenario, $p_{BF}$ and $p_{MA}$ are substituted by sets of prompts where each entry represents a possible bona-fide or morphing attack description, respectively. Let these sets be defined as $P_{BF} = \{p_{BF_1},\,p_{BF_2}\,,  ...,\,p_{BF_N}\}$ and $P_{MA} = \{p_{MA_1},\,p_{MA_2}\,,  ...,\,p_{MA_N}\}$, respectively. Each of the entries of $P_{BF}$ and $P_{MA}$ is individually fed to the text encoder, generating its embedding in the textual feature space. The final embedding representations used for each label ($e_{BF}$ and $e_{MA}$) can then be obtained by averaging the contributions of all the embeddings belonging to the corresponding set:

\begin{equation}
e_{BF} = \frac{1}{N}\sum_{j=1}^N E_t \,(p_{BF_j}), \quad p_{BF_j}\in P_{BF}
\end{equation}

\begin{equation}
e_{MA} = \frac{1}{N}\sum_{j=1}^N E_t \,(p_{MA_j}), \quad p_{MA_j}\in P_{MA}
\end{equation}

Both text embeddings are normalized before being compared with $e_i$ to predict the input sample's class (Equation \ref{eq:predict_label}) to ensure that the embeddings being compared lay on top of a unit hypersphere of the same dimensionality. Note that the usage of multiple prompts per class does not imply additional computational costs when compared with the single text prompt scenario, as the final textual embedding representations $e_{BF}$ and $e_{MA}$ can be pre-computed once and utilized during inference to match with the image embedding.

\begin{table}[]
    \centering
        \caption{Lists with the three sets of prompts representing characteristics linked to face images used in this work: identity, presentation and appearance. For each possible label, the ``\{\}'' field is substituted by its corresponding ISO/IEC 20059 compliant definition (``face image morphing attack'' and ``bona-fide presentation'').}
    \begin{tabular}{c|c|c}
         \textbf{Identity} & \textbf{Presentation} & \textbf{Appearance} \\ \hline
         male \{\}. & frontal \{\}. & bearded \{\}. \\
         female \{\}. & profile \{\}. & moustached \{\}. \\
         young \{\}. & tilted \{\}. & smiling \{\}. \\
         elderly \{\}. & rotated \{\}. & frowning \{\}. \\
         child \{\}. & upward \{\}. & eyeglasses \{\}. \\
         adult \{\}. & downward \{\}. & sunglasses \{\}. \\
         asian \{\}. & sideways \{\}. & wrinkled \{\}. \\
         black \{\}. & leftward \{\}. & balding \{\}. \\
         white \{\}. & rightward \{\}. & occluded \{\}. \\
         latino \{\}. & angled \{\}. & scarred \{\}. \\
         middle eastern \{\}. & inclined \{\}. & pierced \{\}. \\
         indian \{\}. & declined \{\}. & tanned \{\}. \\
         blonde \{\}. & oblique \{\}. & pale \{\}. \\
         brunette \{\}. & twisted \{\}. & makeup \{\}. \\
         redhead \{\}. & turned \{\}. & freckled \{\}. \\
         tall \{\}. & slanted \{\}. & chubby-cheeked \{\}. \\
         short \{\}. & offcenter \{\}. & sweaty \{\}. \\
         thin \{\}. & misaligned \{\}. & dirty \{\}. \\
         obese \{\}. & skewed \{\}. & blinking \{\}. \\
         teen \{\}. & asymmetric \{\}. & tearful \{\}. 
    \end{tabular}

    \label{tab:prompts}
\end{table}           
            
\section{Experimental Setup}
This section presents the experimental setups followed in the paper.
\subsection{Model Architecture}

CLIP \cite{radford2021learning} released four different models with two architectures: base and large. CLIP base architecture has 86M parameters and is available in 2 variants with different patch sizes. CLIP lage contains 0.3 billion parameters and also includes two variants, one of which is pre-trained at a higher resolution for one additional epoch \cite{touvron2019fixing}. The zero-shot MAD performance of these architectures has been assessed by a recent study \cite{caldeira2025madation} that revealed CLIP large architecture's superiority by a significant margin of 12.73 percentage points in terms of average EER across MAD22 \cite{huber2022syn} and its extension MorDIFF \cite{damer2023mordiff}. The higher zero-shot MAD capacity of the large architecture ViT-L is justified by its higher number of parameters, which allows it to learn a more complete set of features and thus perform better in a wider variety of tasks without access to extra knowledge (zero-shot learning) \cite{caldeira2025madation}. Taking these results into consideration, we selected CLIP large trained without high-resolution images as the FM architecture used in this work. This architecture is from now on referred to as ViT-L.

\vspace{-2mm}
\subsection{Text Prompts}

In this work, we evaluate CLIP's zero-shot learning performance in the MAD task when single and multiple prompts are used to describe each classification label. When a single prompt is used per class, we follow the textual descriptions proposed in \cite{caldeira2025madation}, to provide directly comparable results while complying with the ISO/IEC 20059 standard \cite{ISO20059}. However, differently from \cite{caldeira2025madation}, we propose to add a dot to each of the suggested textual prompts to comply with the settings followed during CLIP's training process \cite{radford2021learning}. Hence, the single prompt per class scenario utilizes two possible prompts to describe the analyzed samples: ``face image morphing attack.'' and ``bona-fide presentation.''. The multiple prompt scenario requires a more careful design of the textual inputs, as it delves into more detailed image specificities instead of focusing solely on its possible labels. To provide a thorough investigation of the usefulness of different image attributes in boosting CLIP's zero-shot performance, we considered three attribute lists representing characteristics linked to face images: identity (ID), presentation (Pr) and appearance (Ap). The list of prompts for these three categories are listed in Table \ref{tab:prompts}. For each possible label, the $\{\}$ field is substituted by the correspondent ISO/IEC 20059 compliant definition, as described above. For morphing attack samples, for example, this results in the following ID prompt list: [``male face image morphing attack.'', ``female face image morphing attack.'', ..., ``teen face image morphing attack.'']. Note that all prompts used in the multiple prompt scenario also include a dot in the end, following CLIP's training settings \cite{radford2021learning}. The different lists of prompts are also combined to verify their grouped contribution to zero-shot evaluation, resulting in four extra evaluation settings: ID+Pr, ID+Ap, Pr+Ap and All.

\vspace{-2mm}
\subsection{Image Pre-Processing}

Before being evaluated by the FM, each sample was cropped following \cite{damer2023mordiff} and then resized to 224 × 224 pixels and normalized following the same setting used during CLIP's training ($\mu=$[0.48145466, 0.4578275, 0.40821073], $\sigma=$[0.26862954, 0.26130258, 0.27577711]) \cite{radford2021learning}. These pre-processing steps ensure that the images fed to CLIP comply with the image resolution and normalization settings originally used to train this FM \cite{radford2021learning}.

\subsection{Datasets}

To evaluate the zero-shot learning capacity of CLIP and ensure consistent benchmarking and comparison with earlier research \cite{damer2023mordiff, huber2022syn, caldeira2025madation}, MAD22 \cite{huber2022syn} and its extension MorDIFF \cite{damer2023mordiff} were selected as evaluation datasets. These benchmarks are based on the Face Research Lab London (FRLL)
dataset \cite{DeBruine2021}. They use the same set of 204 bona-fide images and use distinct morphing techniques to create morphing attacks from the same pairs of bona-fide samples. MAD22 includes five sets of morphed samples; two of them were generated by GAN-based representation-level methods (MIPGAN I and II \cite{DBLP:journals/tbbis/ZhangVRRDB21}) while the remaining three derive from image-level techniques (FaceMorpher, OpenCV \cite{openCVmorph}, and Webmorph). MorDIFF's morphing samples were generated with a diffusion autoencoder \cite{DBLP:conf/cvpr/PreechakulCWS22}. 

\subsection{Evaluation Metrics}

The metrics used to perform MAD evaluation in this study were chosen to allow for consistent benchmarking \cite{damer2023mordiff, huber2022syn, caldeira2025madation} while ensuring conformity with the ISO/IEC 30107-3 \cite{ISO301073} standard. These metrics include the Bona-fide Presentation Classification Error Rate (BPCER) and the Attack Presentation Classification Error Rate (APCER), which measure the proportion of bona-fide images misclassified as attack samples and the proportion of attacks misclassified as bona-fide samples, respectively. To cover different operational points and present comparative results, we report both the APCER at fixed BPCER values and the BPCER at fixed APCER values, evaluated at values of 1\%, 10\%, and 20\%. We further report the detection Equal Error Rate (EER), which provides a succinct indicator of the overall performance balance of the system as it corresponds to the error rate at the operating point where the BPCER and APCER are equal.

\subsection{Explainability}

To provide a comprehensive analysis of the performed experiments, we  analyze the relation between different text embeddings and the input images,  supporting our quantitative results with visual explanations. To that end, the similarity score between $e_i$ and a text embedding (obtained by passing single or multiple text prompts to CLIP's text encoder) is backpropagated through the image encoder. This results in a text-conditioned image heatmap that highlights the image regions that are more responsible for the similarity to the analyzed textual embedding ($e_{BF}$ or $e_{MA}$). We start by comparing how the samples from each dataset are activated by single textual prompts with bona-fide or morphing descriptions, as it is expected that they generate distinct heatmaps. We further analyze the impact that using different sets of multiple prompts has on the activations, to verify whether distinct descriptions contribute differently to the final activations.            
            
\section{Results and Discussion}

This section presents a detailed analysis of the results obtained in this work. We start by assessing the importance of complying to the training settings followed during the multimodal FM training process to achieve optimal zero-shot performance. Then, we explore the potential of using multiple textual prompts per label to increase zero-shot learning performance. Finally, we provide an explainability evaluation that showcases CLIP's focus shift when analyzing samples with distinct labels, highlighting its capacity to distinguish bona-fide samples from morphing attacks without fine-tuning to the MAD task.

\vspace{-2mm}
\subsection{The Power of Compliance}

Table \ref{tab:dot} presents CLIP's zero-shot learning performance when using a single prompt to represent each class. The two first sections present the results of MADation (TI) \cite{caldeira2025madation} and our implementation of TI (TI w/o Dot), which used ``face image morphing attack'' and ``bona-fide presentation'' as input text prompts, based on the ISO/IEC 20059 standard. While MADation (TI) normalizes the input samples with a mean and standard deviation of 0.5 on all dimensions, our modified version, TI w/o Dot, defines the normalization hyperparameters as those used during CLIP training ($\mu=$[0.48145466, 0.4578275, 0.40821073], $\sigma=$[0.26862954, 0.26130258, 0.27577711]). The last section, TI-Dot, follows the same sample normalization as TI w/o Dot but further complies with CLIP's training settings by adding the dot character (``.'') in the end of both text prompts.

The analysis of the results shows that TI w/o Dot surpassed MADation's TI \cite{caldeira2025madation} implementation across all evaluated benchmarks and metrics. Although the normalization setting used by TI ($\mu=$[0.5, 0.5, 0.5], $\sigma=$[0.5, 0.5, 0.5]) is commonly followed in MAD systems training and evaluation, this strategy is suboptimal when evaluating CLIP's zero-shot performance, as this FM learned to classify images normalized with a distinct distribution. These findings highlight the importance of correctly adapting the input images' pre-processing depending on the model used to classify them. 

When comparing TI w/o Dot and TI-Dot, the superiority of the latter method is showcased by its decreased average error across 6 of the 7 analyzed metrics. In particular, adding a single dot character (``.'') to the input text prompts fed to CLIP reduced the MAD EER by 1.46 percentage points. These results support the previously withdrawn conclusions, highlighting the importance of correctly following the settings used to train FM's when using them for zero-shot evaluation on domain-specific tasks such as MAD. Taking this into consideration, all the remaining experiments presented in this paper follow the same normalization settings as TI-Dot and use input text prompts that include the dot character.

\begin{table}[tbh!]
\centering
\caption{Evaluation results for CLIP ViT-L using different normalization settings and prompt design structures. The best result achieved for each metric in each test dataset is highlighted in bold.}
\resizebox{0.47\textwidth}{!}
{
\begin{tabular}{|c|c|c|ccc|ccc|}
\hline
\multirow{2}{*}{Method}                                    & \multirow{2}{*}{Test data} & \multirow{2}{*}{EER (\%)} & \multicolumn{3}{c|}{APCER (\%) @ BPCER (\%)}                                                                          & \multicolumn{3}{c|}{BPCER (\%) @ APCER (\%)}                                                                          \\ \cline{4-9} 
\multicolumn{1}{|c|}{}                                                           &                            &                           & \multicolumn{1}{c|}{1.00}                    & \multicolumn{1}{c|}{10.00}                   & 20.00                   & \multicolumn{1}{c|}{1.00}                    & \multicolumn{1}{c|}{10.00}                   & 20.00                   \\ \hline
  \multirow{8}{*}{TI \cite{caldeira2025madation}}                & FaceMorph                  & 44.60                     & \multicolumn{1}{c|}{98.40}                   & \multicolumn{1}{c|}{79.70}                   & 63.60                   & \multicolumn{1}{c|}{99.02}                   & \multicolumn{1}{c|}{87.25}                   & 76.96                   \\
                                 & MIPGAN\_I                  & 18.90            & \multicolumn{1}{c|}{{ 71.80}}             & \multicolumn{1}{c|}{{ 32.20}}             & 17.80          & \multicolumn{1}{c|}{69.61}          & \multicolumn{1}{c|}{33.82}          & { 18.14}             \\
                              & MIPGAN\_II                 & { 12.80}               & \multicolumn{1}{c|}{{ 56.70}}             & \multicolumn{1}{c|}{{ 17.00}}             & 8.90              & \multicolumn{1}{c|}{59.31}          & \multicolumn{1}{c|}{{ 17.16}}             & { 8.33}              \\
                                & OpenCV                     & 35.47                     & \multicolumn{1}{c|}{96.24}                   & \multicolumn{1}{c|}{77.54}                   & 63.11                   & \multicolumn{1}{c|}{96.08}                   & \multicolumn{1}{c|}{73.53}                   & 55.39                   \\
                               & WebMorph                   & 25.20                     & \multicolumn{1}{c|}{94.80}                   & \multicolumn{1}{c|}{52.00}                   & 30.20                   & \multicolumn{1}{c|}{87.75}                   & \multicolumn{1}{c|}{50.98}                   & 32.35                   \\
                                & MorDIFF                    & 42.60                     & \multicolumn{1}{c|}{97.80}                   & \multicolumn{1}{c|}{79.60}                   & 69.50                   & \multicolumn{1}{c|}{97.06}                   & \multicolumn{1}{c|}{83.33}                   & 68.63                   \\ \cline{2-9} 
                                 & \textit{Average}           & \textit{29.93}            & \multicolumn{1}{c|}{\textit{85.96}}          & \multicolumn{1}{c|}{\textit{56.34}}          & \textit{42.19}          & \multicolumn{1}{c|}{\textit{84.81}}          & \multicolumn{1}{c|}{\textit{57.68}}          & \textit{43.30}          \\ \cline{2-9} 
                                  & \textit{Worst}             & \textit{44.60}            & \multicolumn{1}{c|}{\textit{98.40}}          & \multicolumn{1}{c|}{\textit{79.70}}          & \textit{69.50}          & \multicolumn{1}{c|}{\textit{99.02}}          & \multicolumn{1}{c|}{\textit{87.25}}          & \textit{76.96}          \\ \cline{1-9} 
  \multirow{8}{*}{TI w/o Dot (ours)}                & FaceMorph                  &         \textbf{17.70} & \multicolumn{1}{c|}{\textbf{49.50}} & \multicolumn{1}{c|}{\textbf{22.90}} & \multicolumn{1}{c|}{17.30} & \multicolumn{1}{c|}{\textbf{83.82}} & \multicolumn{1}{c|}{35.78} &    \multicolumn{1}{c|}{\textbf{13.73}}    \\
                                & MIPGAN\_I                  &     6.90	& \multicolumn{1}{c|}{\textbf{24.30}} & \multicolumn{1}{c|}{5.70} & \multicolumn{1}{c|}{3.40} & \multicolumn{1}{c|}{35.78} & \multicolumn{1}{c|}{4.41} &	\multicolumn{1}{c|}{1.96}						             \\
                            & MIPGAN\_II                 &     \textbf{3.40}     & \multicolumn{1}{c|}{\textbf{10.51}} & \multicolumn{1}{c|}{\textbf{1.10}} & \multicolumn{1}{c|}{0.70} & \multicolumn{1}{c|}{16.18} & \multicolumn{1}{c|}{\textbf{1.47}} &  \multicolumn{1}{c|}{\textbf{0.49}} \\
                          & OpenCV                     &    16.26  & \multicolumn{1}{c|}{\textbf{59.25}} & \multicolumn{1}{c|}{\textbf{20.73}} & \multicolumn{1}{c|}{14.63} & \multicolumn{1}{c|}{78.92} & \multicolumn{1}{c|}{28.43} &  \multicolumn{1}{c|}{\textbf{10.29}} \\
                              &   WebMorph                   &     \textbf{17.60}   & \multicolumn{1}{c|}{\textbf{62.40}} & \multicolumn{1}{c|}{\textbf{24.00}} & \multicolumn{1}{c|}{\textbf{16.40}} & \multicolumn{1}{c|}{\textbf{67.65}} & \multicolumn{1}{c|}{\textbf{28.43}} & \multicolumn{1}{c|}{\textbf{14.71}}       \\
                            & MorDIFF                    &   32.90  & \multicolumn{1}{c|}{\textbf{93.90}} & \multicolumn{1}{c|}{62.30} & \multicolumn{1}{c|}{49.70} & \multicolumn{1}{c|}{94.12} & \multicolumn{1}{c|}{65.69} & \multicolumn{1}{c|}{48.04} \\ \cline{2-9} 
                & \textit{Average}           & 15.79 & \multicolumn{1}{c|}{\textbf{49.98}} & \multicolumn{1}{c|}{22.79} & \multicolumn{1}{c|}{17.02} & \multicolumn{1}{c|}{62.75} & \multicolumn{1}{c|}{27.37} & \multicolumn{1}{c|}{14.87} \\ \cline{2-9}
                          &   \textit{Worst}             &   32.90  & \multicolumn{1}{c|}{\textbf{93.90}} & \multicolumn{1}{c|}{62.30} & \multicolumn{1}{c|}{49.70} & \multicolumn{1}{c|}{94.12} & \multicolumn{1}{c|}{65.69} & \multicolumn{1}{c|}{48.04} \\ \cline{1-9} 
 \multirow{8}{*}{TI-Dot (ours)}                & FaceMorph                  &          18.10    & \multicolumn{1}{c|}{61.20} & \multicolumn{1}{c|}{25.10} & \multicolumn{1}{c|}{\textbf{16.40}} & \multicolumn{1}{c|}{88.73} & \multicolumn{1}{c|}{\textbf{34.31}} &  \multicolumn{1}{c|}{17.65}  \\
                      & MIPGAN\_I                  &          \textbf{5.40}   & \multicolumn{1}{c|}{26.50} & \multicolumn{1}{c|}{\textbf{4.50}} & \multicolumn{1}{c|}{\textbf{1.60}} & \multicolumn{1}{c|}{\textbf{24.02}} & \multicolumn{1}{c|}{\textbf{3.43}} & \multicolumn{1}{c|}{\textbf{1.47}}   \\
                             & MIPGAN\_II                 &       3.50    & \multicolumn{1}{c|}{13.41} & \multicolumn{1}{c|}{1.20} & \multicolumn{1}{c|}{\textbf{0.20}} & \multicolumn{1}{c|}{\textbf{10.78}} & \multicolumn{1}{c|}{\textbf{1.47}} & \multicolumn{1}{c|}{\textbf{0.49}}   \\
                    & OpenCV                     &  \textbf{16.06}    & \multicolumn{1}{c|}{66.97} & \multicolumn{1}{c|}{21.14} & \multicolumn{1}{c|}{\textbf{10.98}} & \multicolumn{1}{c|}{\textbf{67.16}} & \multicolumn{1}{c|}{\textbf{21.57}} & \multicolumn{1}{c|}{10.78} \\
                  &   WebMorph                   & 18.40  & \multicolumn{1}{c|}{75.60} & \multicolumn{1}{c|}{30.60} & \multicolumn{1}{c|}{17.40} & \multicolumn{1}{c|}{77.45} & \multicolumn{1}{c|}{32.35} &  \multicolumn{1}{c|}{19.12}       \\
                              & MorDIFF                    &   \textbf{24.50} & \multicolumn{1}{c|}{94.70} & \multicolumn{1}{c|}{\textbf{52.20}} & \multicolumn{1}{c|}{\textbf{30.10}} & \multicolumn{1}{c|}{\textbf{91.18}} & \multicolumn{1}{c|}{\textbf{51.96}} & \multicolumn{1}{c|}{\textbf{29.90}} \\ \cline{2-9} 
                        & \textit{Average}           &  \textbf{14.33} & \multicolumn{1}{c|}{56.40} & \multicolumn{1}{c|}{\textbf{22.46}} & \multicolumn{1}{c|}{\textbf{12.78}} & \multicolumn{1}{c|}{\textbf{59.89}} & \multicolumn{1}{c|}{\textbf{24.18}} & \multicolumn{1}{c|}{\textbf{13.24}} \\ \cline{2-9}
                    &   \textit{Worst}             &   \textbf{24.50}   & \multicolumn{1}{c|}{94.70} & \multicolumn{1}{c|}{\textbf{52.20}} & \multicolumn{1}{c|}{\textbf{30.10}} & \multicolumn{1}{c|}{\textbf{91.18}} & \multicolumn{1}{c|}{\textbf{51.96}} &  \multicolumn{1}{c|}{\textbf{29.90}} \\ \hline
\end{tabular}
} \label{tab:dot}
\end{table}

\begin{table}[tbh!]
\centering
\caption{Evaluation results for CLIP ViT-L using a single prompt per class (TI-Dot) and multiple prompts per class. The sets of prompts used for multiple text prompt zero-shot evaluation provide more detailed descriptions than TI-Dot, highlighting identity (ID), presentation (Pr) or appearance (Ap) characteristics, as well as mixtures of these categories (ID+Pr, ID+Ap, Pr+Ap and All). The best result achieved for each metric in each test dataset is highlighted in bold.}
\resizebox{0.47\textwidth}{!}
{
\begin{tabular}{|c|c|c|ccc|ccc|}
\hline
\multicolumn{1}{|c|}{\multirow{2}{*}{Method}}                                    & \multirow{2}{*}{Test data} & \multirow{2}{*}{EER (\%)} & \multicolumn{3}{c|}{APCER (\%) @ BPCER (\%)}                                                                          & \multicolumn{3}{c|}{BPCER (\%) @ APCER (\%)}                                                                          \\ \cline{4-9} 
\multicolumn{1}{|c|}{}                                                           &                            &                           & \multicolumn{1}{c|}{1.00}                    & \multicolumn{1}{c|}{10.00}                   & 20.00                   & \multicolumn{1}{c|}{1.00}                    & \multicolumn{1}{c|}{10.00}                   & 20.00                   \\ \hline
  \multirow{8}{*}{TI-Dot}                & FaceMorph                  &          18.10    & \multicolumn{1}{c|}{61.20} & \multicolumn{1}{c|}{25.10} & \multicolumn{1}{c|}{16.40} & \multicolumn{1}{c|}{88.73} & \multicolumn{1}{c|}{34.31} &  \multicolumn{1}{c|}{17.65}  \\
                          & MIPGAN\_I                  &          5.40    & \multicolumn{1}{c|}{26.50} & \multicolumn{1}{c|}{4.50} & \multicolumn{1}{c|}{1.60} & \multicolumn{1}{c|}{24.02} & \multicolumn{1}{c|}{3.43} & \multicolumn{1}{c|}{1.47}   \\
                              & MIPGAN\_II                 &       3.50    & \multicolumn{1}{c|}{13.41} & \multicolumn{1}{c|}{1.20} & \multicolumn{1}{c|}{0.20} & \multicolumn{1}{c|}{10.78} & \multicolumn{1}{c|}{1.47} &  \multicolumn{1}{c|}{\textbf{0.49}}  \\
                          & OpenCV                     &  16.06    & \multicolumn{1}{c|}{66.97} & \multicolumn{1}{c|}{21.14} & \multicolumn{1}{c|}{10.98} & \multicolumn{1}{c|}{67.16} & \multicolumn{1}{c|}{21.57} & \multicolumn{1}{c|}{10.78} \\
                               &   WebMorph                   & \textbf{18.40}  & \multicolumn{1}{c|}{75.60} & \multicolumn{1}{c|}{\textbf{30.60}} & \multicolumn{1}{c|}{\textbf{17.40}} & \multicolumn{1}{c|}{\textbf{77.45}} & \multicolumn{1}{c|}{\textbf{32.35}} &  \multicolumn{1}{c|}{\textbf{19.12}}       \\
                             & MorDIFF                    &   24.50 & \multicolumn{1}{c|}{94.70} & \multicolumn{1}{c|}{52.20} & \multicolumn{1}{c|}{30.10} & \multicolumn{1}{c|}{91.18} & \multicolumn{1}{c|}{51.96} & \multicolumn{1}{c|}{29.90} \\ \cline{2-9} 
                         & \textit{Average}           &  14.33 & \multicolumn{1}{c|}{56.40} & \multicolumn{1}{c|}{22.46} & \multicolumn{1}{c|}{12.78} & \multicolumn{1}{c|}{59.89} & \multicolumn{1}{c|}{24.18} & \multicolumn{1}{c|}{13.24} \\ \cline{2-9} 
                   &   \textit{Worst}             &   24.50   & \multicolumn{1}{c|}{94.70} & \multicolumn{1}{c|}{52.20} & \multicolumn{1}{c|}{30.10} & \multicolumn{1}{c|}{91.18} & \multicolumn{1}{c|}{51.96} &  \multicolumn{1}{c|}{29.90} \\ \cline{1-9}

 \multirow{8}{*}{ID}                & FaceMorph                  &       19.90          & \multicolumn{1}{c|}{62.80}                    & \multicolumn{1}{c|}{31.60}                    &        19.90         & \multicolumn{1}{c|}{84.80}                   & \multicolumn{1}{c|}{38.24}                   &          20.10          \\
               & MIPGAN\_I                  &        7.00        & \multicolumn{1}{c|}{21.30}                    & \multicolumn{1}{c|}{5.60}                    &        2.70         & \multicolumn{1}{c|}{25.98}                   & \multicolumn{1}{c|}{5.39}                   &          1.47          \\
                      & MIPGAN\_II                 &      5.01         & \multicolumn{1}{c|}{13.61}                    & \multicolumn{1}{c|}{2.10}                    &         0.80        & \multicolumn{1}{c|}{19.12}                   & \multicolumn{1}{c|}{3.43}                   &    0.49    \\
                           & OpenCV                     &       16.06            & \multicolumn{1}{c|}{55.69}                    & \multicolumn{1}{c|}{21.65}                    &         11.69        & \multicolumn{1}{c|}{67.65}                   & \multicolumn{1}{c|}{22.06}                   &     12.75    \\
                           & WebMorph                   &      22.00         & \multicolumn{1}{c|}{\textbf{67.20}}                    & \multicolumn{1}{c|}{36.20}                    &      24.40           & \multicolumn{1}{c|}{85.78}                   & \multicolumn{1}{c|}{38.24}                   &   24.51    \\
                       & MorDIFF                     &       21.40          & \multicolumn{1}{c|}{82.30}                    & \multicolumn{1}{c|}{43.20}                    &        27.00         & \multicolumn{1}{c|}{86.27}                   & \multicolumn{1}{c|}{37.75}                   &         24.02           \\ \cline{2-9} 
                            & \textit{Average}           &       15.23          & \multicolumn{1}{c|}{50.48}                    & \multicolumn{1}{c|}{23.39}                    &        14.43         & \multicolumn{1}{c|}{61.60}                   & \multicolumn{1}{c|}{24.19}                   &      13.89     \\ \cline{2-9} 
                            & \textit{Worst}             &        \textbf{22.00}         & \multicolumn{1}{c|}{82.30}                    & \multicolumn{1}{c|}{43.20}                    &        27.00         & \multicolumn{1}{c|}{86.27}                   & \multicolumn{1}{c|}{\textbf{38.24}}                   &     \textbf{24.51}      \\ \cline{1-9}

 \multirow{8}{*}{Pr}                & FaceMorph                  &        14.00         & \multicolumn{1}{c|}{\textbf{53.60}}                    & \multicolumn{1}{c|}{\textbf{17.40}}                    &     10.60            & \multicolumn{1}{c|}{71.57}                   & \multicolumn{1}{c|}{21.08}                   &      \textbf{8.33}    \\
\multirow{8}{*}{}                & MIPGAN\_I                  &        5.40         & \multicolumn{1}{c|}{18.90}                    & \multicolumn{1}{c|}{3.10}                    &     0.80            & \multicolumn{1}{c|}{18.63}                   & \multicolumn{1}{c|}{\textbf{2.94}}                   &     0.98      \\
 \multirow{8}{*}{}                & MIPGAN\_II                 &        3.40         & \multicolumn{1}{c|}{9.91}                    & \multicolumn{1}{c|}{\textbf{0.90}}                    &         0.20        & \multicolumn{1}{c|}{\textbf{8.33}}                   & \multicolumn{1}{c|}{0.98}                   &      \textbf{0.49}     \\
                      & OpenCV                     &         13.21        & \multicolumn{1}{c|}{57.62}                    & \multicolumn{1}{c|}{\textbf{15.85}}                    &         10.06        & \multicolumn{1}{c|}{\textbf{56.37}}                   & \multicolumn{1}{c|}{20.10}                   &     \textbf{7.35}   \\
                           & WebMorph                    &         25.60        & \multicolumn{1}{c|}{81.60}                    & \multicolumn{1}{c|}{43.60}                    &       30.20          & \multicolumn{1}{c|}{83.82}                 & \multicolumn{1}{c|}{47.06}                   &       33.33     \\
                             & MorDIFF                    &       14.70          & \multicolumn{1}{c|}{73.90}                    & \multicolumn{1}{c|}{20.40}                    &        9.70         & \multicolumn{1}{c|}{72.55}                   & \multicolumn{1}{c|}{19.12}                   &    \textbf{10.29}  \\ \cline{2-9} 
                         & \textit{Average}            &         12.72        & \multicolumn{1}{c|}{49.26}                    & \multicolumn{1}{c|}{\textbf{16.88}}                    &      10.26           & \multicolumn{1}{c|}{51.88}                   & \multicolumn{1}{c|}{18.55}                   &    \textbf{10.13}     \\ \cline{2-9} 
                               & \textit{Worst}              &        25.60         & \multicolumn{1}{c|}{81.60}                    & \multicolumn{1}{c|}{43.60}                    &       30.20          & \multicolumn{1}{c|}{\textbf{83.82}}                   & \multicolumn{1}{c|}{47.06}                   &      33.33    \\ \cline{1-9}

  \multirow{8}{*}{Ap}                & FaceMorph                   &       15.50          & \multicolumn{1}{c|}{62.70}                    & \multicolumn{1}{c|}{23.70}                    &       \textbf{9.10}          & \multicolumn{1}{c|}{\textbf{66.67}}                   & \multicolumn{1}{c|}{\textbf{18.63}}                   &      11.76     \\
                              & MIPGAN\_I                  &        6.40         & \multicolumn{1}{c|}{19.00}                    & \multicolumn{1}{c|}{\textbf{2.50}}                    &       0.60          & \multicolumn{1}{c|}{14.71}                   & \multicolumn{1}{c|}{3.43}                   &   0.98    \\
                                & MIPGAN\_II                  &        \textbf{3.00}         & \multicolumn{1}{c|}{12.21}                    & \multicolumn{1}{c|}{1.00}                    &         \textbf{0.10}        & \multicolumn{1}{c|}{9.80}                   & \multicolumn{1}{c|}{1.47}                   &    \textbf{0.49}    \\
                              & OpenCV                      &       13.52          & \multicolumn{1}{c|}{62.30}                    & \multicolumn{1}{c|}{20.33}                    &       \textbf{8.13}         & \multicolumn{1}{c|}{57.84}                   & \multicolumn{1}{c|}{17.65}                   &     9.80     \\
                           & WebMorph                   &        24.60         & \multicolumn{1}{c|}{76.80}                    & \multicolumn{1}{c|}{46.60}                    &         26.80        & \multicolumn{1}{c|}{92.16}                   & \multicolumn{1}{c|}{57.35}                   &       31.86     \\
                       & MorDIFF                   &       15.00          & \multicolumn{1}{c|}{78.90}                    & \multicolumn{1}{c|}{30.20}                    &       \textbf{8.20}          & \multicolumn{1}{c|}{\textbf{60.78}}                   & \multicolumn{1}{c|}{\textbf{17.65}}                   &    12.75     \\ \cline{2-9} 
                    & \textit{Average}           &        13.00         & \multicolumn{1}{c|}{51.99}                    & \multicolumn{1}{c|}{20.72}                    &     \textbf{8.82}            & \multicolumn{1}{c|}{\textbf{50.33}}                   & \multicolumn{1}{c|}{19.36}                   &    11.27     \\ \cline{2-9} 
                           & \textit{Worst}             &        24.60         & \multicolumn{1}{c|}{78.90}                    & \multicolumn{1}{c|}{46.60}                    &    26.80       & \multicolumn{1}{c|}{92.16}                   & \multicolumn{1}{c|}{57.35}                   &      31.86      \\ \cline{1-9}

 \multirow{8}{*}{ID+Pr}            & FaceMorph                  &        16.20         & \multicolumn{1}{c|}{58.70}                    & \multicolumn{1}{c|}{23.30}                    &         13.90        & \multicolumn{1}{c|}{79.90}                   & \multicolumn{1}{c|}{27.45}                   &   10.78  \\  
                      & MIPGAN\_I                   &         5.90        & \multicolumn{1}{c|}{19.20}                    & \multicolumn{1}{c|}{3.90}                    &        1.90         & \multicolumn{1}{c|}{23.04}                   & \multicolumn{1}{c|}{3.43}                   &      0.98         \\  
                           & MIPGAN\_II                 &        3.20         & \multicolumn{1}{c|}{11.31}                    & \multicolumn{1}{c|}{1.30}                    &      0.20           & \multicolumn{1}{c|}{10.78}                   & \multicolumn{1}{c|}{1.96}                   &    \textbf{0.49}      \\  
                           & OpenCV                     &        13.62         & \multicolumn{1}{c|}{56.61}                    & \multicolumn{1}{c|}{19.21}                    &       10.57          & \multicolumn{1}{c|}{62.25}                   & \multicolumn{1}{c|}{21.08}                   &       9.31      \\  
                          & WebMorph                   &         23.40        & \multicolumn{1}{c|}{75.60}                    & \multicolumn{1}{c|}{41.40}                    &      27.80        & \multicolumn{1}{c|}{86.76}                   & \multicolumn{1}{c|}{45.59}                   &     27.45      \\  
                & MorDIFF                    &        18.40         & \multicolumn{1}{c|}{79.00}                    & \multicolumn{1}{c|}{\textbf{15.80}}                    &        31.80         & \multicolumn{1}{c|}{81.86}                   & \multicolumn{1}{c|}{27.45}                   &       15.69       \\ \cline{2-9}  
                      & \textit{Average}          &         13.45        & \multicolumn{1}{c|}{50.07}                    & \multicolumn{1}{c|}{17.49}                    &        14.36         & \multicolumn{1}{c|}{57.43}                   & \multicolumn{1}{c|}{21.16}                   &       10.78       \\ \cline{2-9}  
                        & \textit{Worst}             &        23.40         & \multicolumn{1}{c|}{79.00}                    & \multicolumn{1}{c|}{41.40}                    &     31.80       & \multicolumn{1}{c|}{86.76}                   & \multicolumn{1}{c|}{45.59}                   &       27.45       \\ \cline{1-9}

\multirow{8}{*}{ID+Ap}            & FaceMorph                 &        16.10         & \multicolumn{1}{c|}{60.40}                    & \multicolumn{1}{c|}{25.60}                    &       14.30          & \multicolumn{1}{c|}{77.45}                   & \multicolumn{1}{c|}{26.96}                   &     12.75      \\  
                             & MIPGAN\_I                  &        6.50         & \multicolumn{1}{c|}{17.70}                    & \multicolumn{1}{c|}{3.90}                    &       0.80          & \multicolumn{1}{c|}{16.67}                   & \multicolumn{1}{c|}{4.41}                   &      0.98    \\  
                            & MIPGAN\_II                 &        4.60         & \multicolumn{1}{c|}{10.71}                    & \multicolumn{1}{c|}{1.10}                    &        0.20         & \multicolumn{1}{c|}{11.76}                   & \multicolumn{1}{c|}{1.96}                   &     \textbf{0.49}   \\  
                        & OpenCV                      &        13.85         & \multicolumn{1}{c|}{55.59}                    & \multicolumn{1}{c|}{19.51}                    &        10.06         & \multicolumn{1}{c|}{63.24}                   & \multicolumn{1}{c|}{19.61}                   &     9.31     \\  
                             & WebMorph             &         22.20        & \multicolumn{1}{c|}{71.20}                    & \multicolumn{1}{c|}{41.00}                    &         26.00        & \multicolumn{1}{c|}{88.73}                   & \multicolumn{1}{c|}{48.53}                   &      27.94    \\  
                        & MorDIFF                &         17.70        & \multicolumn{1}{c|}{79.10}                    & \multicolumn{1}{c|}{35.00}                    &         16.20        & \multicolumn{1}{c|}{78.43}                   & \multicolumn{1}{c|}{27.94}                   &      15.20     \\ \cline{2-9}  
                         & \textit{Average}           &        13.49         & \multicolumn{1}{c|}{49.12}                    & \multicolumn{1}{c|}{21.02}                    &      11.26           & \multicolumn{1}{c|}{56.05}                   & \multicolumn{1}{c|}{21.57}                   &       11.11     \\ \cline{2-9}  
                        & \textit{Worst}             &         22.20        & \multicolumn{1}{c|}{79.10}                    & \multicolumn{1}{c|}{\textbf{41.00}}                    &     \textbf{26.00}	& \multicolumn{1}{c|}{88.73}                   & \multicolumn{1}{c|}{48.53}                   &        27.94      \\ \cline{1-9}

 \multirow{8}{*}{Pr+Ap}            & FaceMorph                  &       \textbf{12.90}          & \multicolumn{1}{c|}{55.40}                    & \multicolumn{1}{c|}{19.80}                    &        9.50         & \multicolumn{1}{c|}{70.59}                   & \multicolumn{1}{c|}{19.61}                   &      9.80      \\  
                         & MIPGAN\_I                  &       \textbf{4.50}          & \multicolumn{1}{c|}{\textbf{16.70}}                    & \multicolumn{1}{c|}{3.20}                    &       \textbf{0.50}          & \multicolumn{1}{c|}{\textbf{13.73}}                   & \multicolumn{1}{c|}{3.43}                   &         \textbf{0.49}         \\  
                             & MIPGAN\_II                &        3.60         & \multicolumn{1}{c|}{\textbf{9.81}}                    & \multicolumn{1}{c|}{\textbf{0.90}}                    &      \textbf{0.10}           & \multicolumn{1}{c|}{9.80}                   & \multicolumn{1}{c|}{\textbf{0.49}}                   &        \textbf{0.49}        \\  
                             & OpenCV                   &        12.80         & \multicolumn{1}{c|}{55.79}                    & \multicolumn{1}{c|}{18.70}                    &      9.04           & \multicolumn{1}{c|}{56.86}                   & \multicolumn{1}{c|}{\textbf{16.67}}                   &       7.84          \\  
                              & WebMorph         &         23.60        & \multicolumn{1}{c|}{77.60}                    & \multicolumn{1}{c|}{47.00}                    &        28.80         & \multicolumn{1}{c|}{87.75}                   & \multicolumn{1}{c|}{51.96}                   &           31.86        \\  
                               & MorDIFF       &        \textbf{12.30}         & \multicolumn{1}{c|}{\textbf{73.40}}                    & \multicolumn{1}{c|}{26.80}                    &        8.60         & \multicolumn{1}{c|}{69.61}                   & \multicolumn{1}{c|}{\textbf{17.65}}                   &     11.27     \\ \cline{2-9}  
                              & \textit{Average}         &       \textbf{11.62}          & \multicolumn{1}{c|}{\textbf{48.12}}                    & \multicolumn{1}{c|}{19.40}                    &      9.42           & \multicolumn{1}{c|}{51.39}                   & \multicolumn{1}{c|}{\textbf{18.30}}                   &         10.29     \\ \cline{2-9}  
                                & \textit{Worst}            &      23.60           & \multicolumn{1}{c|}{\textbf{77.60}}                    & \multicolumn{1}{c|}{47.00}                    &        28.80         & \multicolumn{1}{c|}{87.75}                   & \multicolumn{1}{c|}{51.96}                   &       31.86   \\ \cline{1-9}

\multirow{8}{*}{All}          & FaceMorph       &        14.90         & \multicolumn{1}{c|}{58.60}                    & \multicolumn{1}{c|}{21.40}                    &       11.80          & \multicolumn{1}{c|}{75.49}                   & \multicolumn{1}{c|}{25.00}                   &     10.78   \\  
                            & MIPGAN\_I     &        5.40         & \multicolumn{1}{c|}{17.10}                    & \multicolumn{1}{c|}{3.80}                    &        0.60         & \multicolumn{1}{c|}{17.16}                   & \multicolumn{1}{c|}{3.43}                   &       \textbf{0.49}     \\  
                           & MIPGAN\_II    &        3.80         & \multicolumn{1}{c|}{10.21}                    & \multicolumn{1}{c|}{10.00}                    &       0.30          & \multicolumn{1}{c|}{10.78}                   & \multicolumn{1}{c|}{0.98}                   &      \textbf{0.49}    \\  
                           & OpenCV    &       \textbf{12.30}         & \multicolumn{1}{c|}{\textbf{55.18}}                    & \multicolumn{1}{c|}{18.50}                    &         9.65        & \multicolumn{1}{c|}{62.25}                   & \multicolumn{1}{c|}{19.61}                   &    8.82     \\  
                             & WebMorph   &        22.80         & \multicolumn{1}{c|}{74.80}                    & \multicolumn{1}{c|}{42.60}                    &        27.20         & \multicolumn{1}{c|}{85.78}                   & \multicolumn{1}{c|}{47.06}                   &    28.92    \\  
                                & MorDIFF   &        15.50         & \multicolumn{1}{c|}{77.60}                    & \multicolumn{1}{c|}{30.40}                    &       12.00          & \multicolumn{1}{c|}{76.47}                   & \multicolumn{1}{c|}{24.02}                   &   13.24   \\ \cline{2-9}  
                             & \textit{Average}       &        12.45         & \multicolumn{1}{c|}{48.92}                    & \multicolumn{1}{c|}{21.12}                    &        10.26         & \multicolumn{1}{c|}{54.66}                   & \multicolumn{1}{c|}{20.02}                   &     10.46  \\ \cline{2-9}  
                             & \textit{Worst}          &        22.80         & \multicolumn{1}{c|}{\textbf{77.60}}                    & \multicolumn{1}{c|}{42.60}                    &         27.20        & \multicolumn{1}{c|}{85.78}                   & \multicolumn{1}{c|}{47.06}                   &     28.92   \\ \hline
\end{tabular}
} \label{tab:multi_prompt}
\end{table}

\vspace{-2mm}
\subsection{Multiple Prompt Aggregation}

Table \ref{tab:multi_prompt} presents CLIP zero-shot learning performance when multiple text prompts' contributions are averaged for each class (morphing attacks and bona-fide). These results are directly compared with the scenario where a single text prompt per label is used (TI-Dot) to quantify the impact of using multiple text prompts per class. It can be seen that from 6 out of the 7 multiple prompt settings surpassed TI-Dot in terms of average EER. In particular, the Pr+Ap setting achieved the best overall performance, reducing the average EER achieved with a single prompt per class by the considerable margin of 2.71 percentage points. 

While it is important to determine which prompt setting leads to better zero-shot performance, it is also relevant to understand how the different sets of prompts contribute to improving the performance achieved with a single text prompt per class. An initial assessment of each individual set of prompts (identity, presentation, appearance) can be done by comparing the performance of settings ID, Pr, and Ap, respectively, with the baseline, TI-Dot. While ID falls behind TI-Dot in 6 out of the 7 averaged metrics, Pr and Ap managed to surpass the single text prompt approach in all 7 metrics, with Pr performing better than Ap in 5 of them. These results suggest that while both presentation and appearance attribute information positively contribute to increasing the performance of the FM zero-shot performance, the incorporation of id-related information is not beneficial. This conclusion is also supported by the fact that ID+Pr shows increased performance in comparison to ID while falling behind Pr in all 7 averaged metrics. Similar conclusions can also be withdrawn when comparing ID+Ap with ID and Ap or All with Pr+Ap. The Pr+Ap setting, on the other hand, manages to surpass Pr and Ap in most of the considered evaluation metrics, suggesting that the presentation and appearance attributes provide information presenting complementary benefits to the FM zero-shot capacity. Overall, Pr+Ap proved to be the best performing approach, surpassing all remaining strategies in 3 out of the 7 average metrics, including the EER. The best average value for the 4 remaining metrics was achieved by either Pr or Ap, supporting the individual contribution of these two settings towards increased zero-shot performance. 

\begin{figure*}[ht!]
    \centering
    \includegraphics[width=0.9\linewidth]{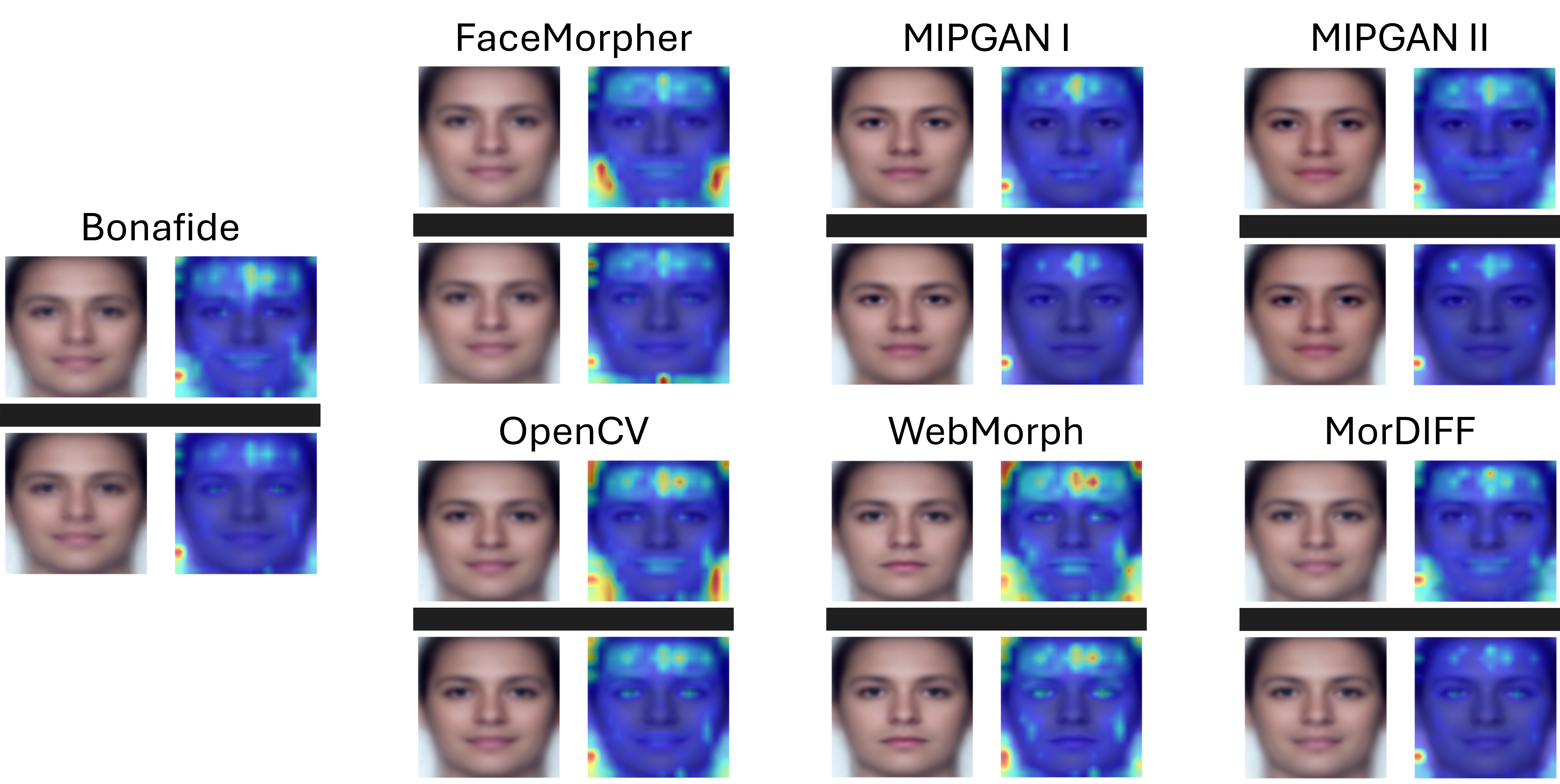}
    \caption{CLIP's average activation heatmaps for different MAD datasets when using a single text prompt to describe each possible class (TI-Dot). Each block represents one dataset, displaying an average representation of all its samples as well as the correspondent average activation heatmap. For each dataset, the first and second line highlight the heatmap associated with the morphing attack and the bona-fide prompt, respectively. Since all the evaluation datasets considered in this work share the same set of bona-fide samples, the average bona-fide activation maps are highlighted separately (first column). The average input sample and heatmap associated with each dataset (second to foruth columns) correspond only to its morphing attack samples.}
    \label{fig:explainability}

\end{figure*}

\begin{figure*}[ht!]
    \centering
    \includegraphics[width=0.9\linewidth]{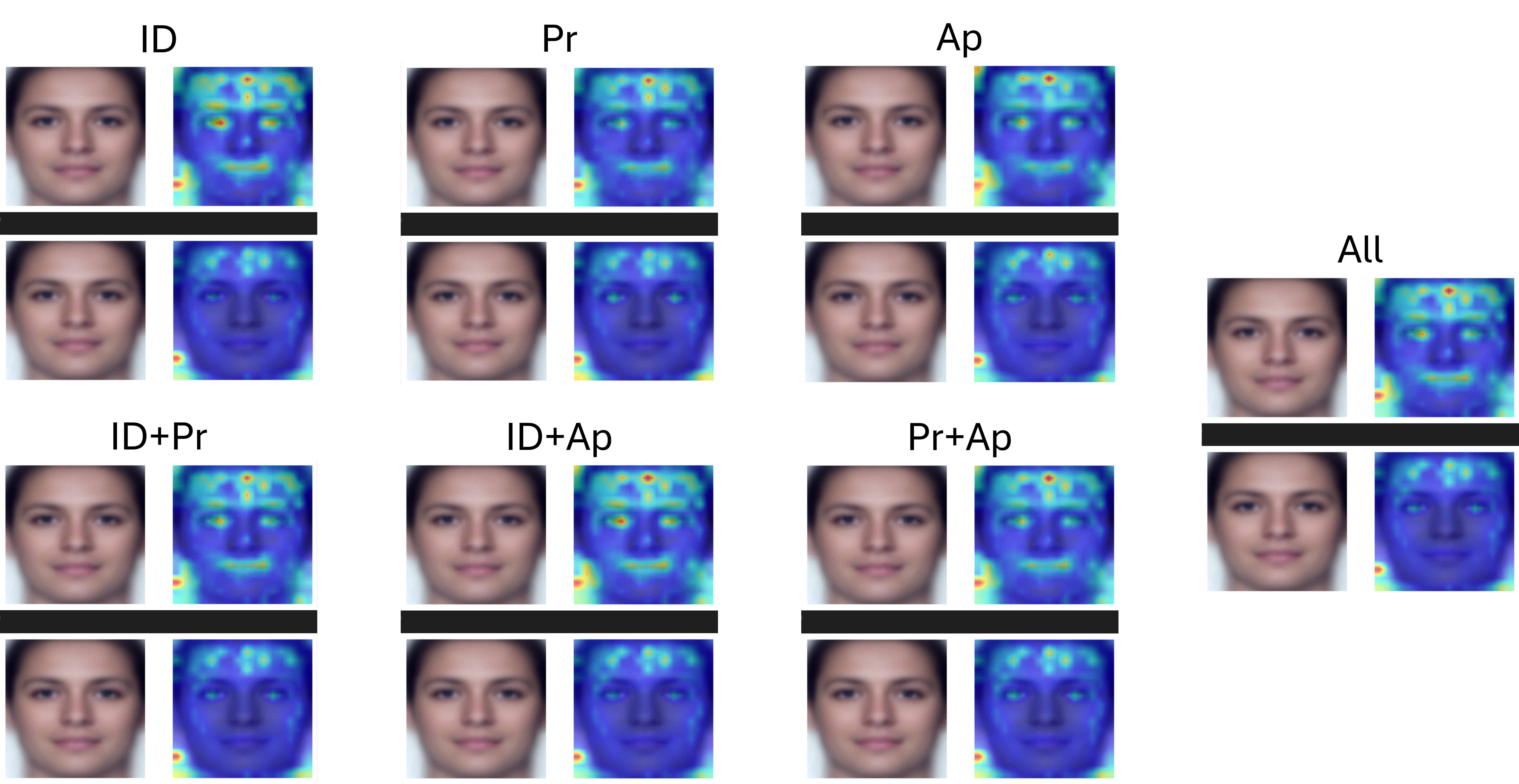}
    \caption{CLIP's average activation heatmaps for the morphing samples on the MorDIFF dataset when using multiple text prompts to describe each possible class (ID, Pr, Ap, ID+Pr, ID+Ap, Pr+Ap, All). For each setting, the first and second line highlight the heatmap associated with the morphing attack and the bona-fide sets of prompts, respectively.}
    \label{fig:mordiff}
\end{figure*}

While this study does not aim to achieve SOTA performance but to provide a comprehensive analysis of zero-shot performance when using different text input prompts and raise awareness towards the importance of efficient prompt engineering and prompt aggregation, the comparison between our MADPromptS strategy and MADation \cite{caldeira2025madation} derives naturally from the previous comparison established with the zero-shot learning results presented in this work (Table \ref{tab:dot}). It is interesting to notice that Pr+Ap surpasses the fine-tuned CLIP model proposed in \cite{caldeira2025madation} by 0.32 percentage points in terms of average EER, without requiring any extra fine-tuning or adaption to the downstream MAD task. The complete assessment of the results provided in this section reveals the importance of selecting appropriate text prompts when using FMs in the zero-shot evaluation setting, highlighting the benefits that can arise from efficient prompt engineer. Hence, this work paves the way towards more efficient prompt engineering and aggregation, providing important insights regarding the importance of correctly exploiting textual prompts to the user's advantage by leveraging FM’s zero-shot learning ability to a fuller extent.

\subsection{Explainability}														

Figure \ref{fig:explainability} shows CLIP's average activation heatmaps of different datasets when using a single text prompt to describe each possible class (TI-Dot). For each dataset, the first and second row highlight the heatmap associated with the morphing attack and the bona-fide prompt, respectively. All the evaluation datasets considered in this work share the same set of bona-fide samples. Hence, the activation maps for the bona-fide samples are highlighted separately, and the average input sample and heatmap associated with each dataset corresponds only to its morphing attack samples. It should be noted that all these datasets use the same pairs of bona-fide samples to create the morphing, which justifies the high similarity between the average input samples shown for each scenario. It is possible to observe that the heatmaps obtained when the morphing prompt is encoded by the text encoder significantly differ from the ones associated with bona-fide text prompts. In particular, the usage of morphing prompts makes CLIP focus its attention in detailed areas of the face, such as the mouth and jawline. These areas are particularly prone to contain artifacts in morphing samples \cite{damer2023mordiff, DBLP:journals/tbbis/ZhangVRRDB21}, highlighting CLIP's capacity to focus on important characteristics that might indicate the presence of a malicious sample without any fine-tuning (zero-shot learning). It is also worth notice that the heatmaps for presented morphing prompts generally show more activation when analyzing morphing samples in comparison with bona-fide images. These results complement the previous analysis regarding the effectiveness of the proposed method, providing a visual and easily interpretable explanation of the obtained results.

Similar heatmaps were also plotted for the different multiple prompt scenarios analyzed in this work (Figure \ref{fig:mordiff}). This plot follows the same presentation logic as Figure \ref{fig:explainability} and focuses on the MorDIFF dataset, for which our ID+Pr multiple prompt strategy surpassed the baseline (TI-Dot) by a larger margin. It can be seen that using different sets of prompts to describe the input samples results in different activation patterns, supporting their distinct influence in CLIP's performance. In particular, it is interesting to observe that combining different sets of prompts in the proposed multiple text prompt zero-shot approach results in activation maps that combine characteristics from all the incorporated sets. As an example, when analyzing the morphing heatmaps of ID, Pr and Ap it is clear that ID results in a stronger activation in the eye region, followed by Ap and finally by Pr. This tendency is kept when analysing joint contributions, with ID+Ap showcasing the strongest activation in the eye region, followed by ID+Pr and, finally, by Pr+Ap. The complementary nature of these activations is also in line with the conclusions withdrawn in the previous section revealing that sets of prompts that contribute to increase the baseline (TI-Dot) performance can generally be combined to boost CLIP's zero-shot MAD capacity even more (Pr+Ap vs Pr and Ap) while combining a set that negatively impacts CLIP performance with other prompt collections reduces the performance achieved with a single benefitial prompt set (ID+Pr vs Pr, for example). 

Overall, it is possible to conclude that using single or multiple prompts per class results in significantly different CLIP activations (Figures \ref{fig:explainability} and \ref{fig:mordiff}), and that CLIP follows distinct attention patterns when using different multiple prompt sets. These conclusions are in line with the results displayed in Table \ref{tab:multi_prompt}, highlighting the importance of effective prompt engineering to take the best possible advantage of FM's zero-shot learning capacity.             
            
\section{Conclusion}
This work offers a comprehensive analysis of the use of multimodal FMs for the critical task of MAD under a zero-shot setting. We demonstrate that careful alignment with the model’s original training conditions, including appropriate textual prompt design, can significantly enhance zero-shot performance without any fine-tuning. Beyond this, we introduce and evaluate a strategy of aggregating multiple carefully designed textual prompts per class, enabling the model to capture more diverse and discriminative cues relevant to distinguishing bona-fide from attack samples. While using a single prompt per class provides only a general description of the target label, employing multiple prompts incorporates more specific characteristics, shifting the model’s attention to a broader range of details. Our experiments show that disjoint sets of prompts exhibit complementary capabilities when combined. Moreover, leveraging prompt aggregation not only enhances zero-shot MAD performance but can even surpass fine-tuned models, highlighting the untapped potential of well-designed textual prompts as a simple yet effective alternative to task-specific fine-tuning. These findings emphasize the importance of prompt design for zero-shot FM predictions, paving the way for more generalizable and scalable MAD solutions in biometric security applications.            

\section*{Acknowledgments}

This research work has been funded by the German Federal Ministry of Education and Research and the Hessian Ministry of Higher Education, Research, Science and the Arts within their joint support of the National Research Center for Applied Cybersecurity ATHENE.

\bibliographystyle{ACM-Reference-Format}
\balance
\bibliography{sample-base}

\end{document}